\documentclass[runningheads]{llncs}
\usepackage{graphicx}
\usepackage{hyperref}
\usepackage{breakurl}
\usepackage{listings}
\usepackage{multirow}
\usepackage{color}
\usepackage{booktabs}  
\usepackage{multirow}  
\usepackage[backend=biber, isbn=false, doi=false, sorting=none]{biblatex}
\usepackage{subfig}
\usepackage{amsmath}
\usepackage[T1]{fontenc} 
\usepackage{xspace}
\usepackage[inline]{enumitem} 
\usepackage{cleveref} 
\usepackage{tablefootnote}

\newcommand{\placeholder}[1]{\texttt{\textless#1\textgreater}}
\newcommand{\verbLU}{\textsc{PlainText}\xspace}
\newcommand{\verbAF}{\textsc{SpatialFormat}\xspace}
\newcommand{\verbAFY}{\textsc{SpatialFormatY}\xspace}
\newcommand{\verbBB}{\textsc{BoundingBox}\xspace}
\newcommand{\verbBBM}{\textsc{BoundingBoxMarkup}\xspace}
\newcommand{\verbCP}{\textsc{CenterPoint}\xspace}
\newcommand{\verbPH}{\textsc{PlainHTML}\xspace}
\newcommand{\noiseNONE}{\texttt{NONE}\xspace}
\newcommand{\noiseBBOXES}{\texttt{TRANSLATE}\xspace}
\newcommand{\noiseSHUFFLE}{\texttt{SHUFFLE}\xspace}
\newcommand{\noiseSEGMENT}{\texttt{NEAREST\_NEIGHBOR}\xspace}

\bibliography{main.bib}

\begin{document}
\title{LAPDoc: Layout-Aware Prompting for Documents\thanks{Currently under review at ICDAR2024.}}

\author{Marcel Lamott\inst{1} \and
Yves-Noel Weweler\inst{2} \and
Adrian Ulges\inst{1} \and
Faisal Shafait\inst{3} \and
Dirk Krechel\inst{1}\orcidID{0000-0003-0984-5918} \and
Darko Obradovic\inst{2}}
\authorrunning{M. Lamott et al.}
%
\institute{
RheinMain University of Applied Sciences, Wiesbaden, Germany\\
\email{\{marcel.lamott\footnote{Corresponding author}, adrian.ulges, dirk.krechel\}@hs-rm.de}
 \and
Insiders Technologies GmbH, Kaiserslautern, Germany\\
\email{\{y.weweler, d.obradovic\}@insiders-technologies.de}
\and
National University of Sciences and Technology, Islamabad, Pakistan\\
\email{faisal.shafait@seecs.edu.pk}}
\maketitle              
\begin{abstract}
Recent advances in training large language models (LLMs) using massive amounts of solely textual data lead to strong generalization across many domains and tasks, including document-specific tasks.
Opposed to that there is a trend to train multi-modal transformer architectures tailored for document understanding that are designed specifically to fuse textual inputs with the corresponding document layout. 
This involves a separate fine-tuning step for which additional training data is required. At present, no document transformers with comparable generalization to LLMs are available
That raises the question which type of model is to be preferred for document understanding tasks.
In this paper we investigate the possibility to use purely text-based LLMs for document-specific tasks by using layout enrichment. We explore drop-in modifications and rule-based methods to enrich purely textual LLM prompts with layout information. In our experiments we investigate the effects on the commercial ChatGPT model and the open-source LLM Solar.
We demonstrate that using our approach both LLMs show improved
performance on various standard document benchmarks. In addition, we study the impact
of noisy OCR and layout errors, as well as the limitations of LLMs when
it comes to utilizing document layout. Our results indicate that layout
enrichment can improve the performance of purely text-based LLMs for
document understanding by up to 15\% compared to just using plain
document text.
In conclusion, this approach should be considered for the best model choice between text-based LLM or multi-modal document transformers.
\keywords{Document Understanding \and Large Language Models \and Layout Understanding \and Prompt Enrichment}
\end{abstract}

\section{Introduction}
In today's business environment, companies face the problem of an ever growing 
amount of digital documents that need to be processed.
The possibilities of smart devices to capture documents has lead to new digital business models that make heavy use of camera captures, while promising high automation.
This induces a dramatic growth in digitized documents of varying quality that need to be processed.
In addition to document types such as invoices, forms, complaints, receipts and notices, other, less standardized types of documents are increasingly important, such as contract documents, business reports or legal texts.

Understanding documents completely necessitates understanding of textual and visual modalities as well as the comprehension of the spatial relations between the document's content elements, 
which guide the reading process and are essential for interpretation~\cite[p.~1]{Borchmann2021DUEED}.
Recently, automated document image understanding has taken strides forward: (i) Larger-scale benchmarks that align with real applications~\cite{Borchmann2021DUEED, li2020tablebank, li2020docbank, xu-etal-2022-xfund} allow for real world evaluation and training. (ii) Self-supervised pre-training tasks -- with which  large amounts of data can be leveraged without the need for hand-crafted annotations -- have led to multi-modal neural models.
These models can either take document images and text (e.g., extracted by OCR) as input~\cite{cao2022gmn, tang2023unifying, luo2023geolayoutlm, wang2023docllm}, or can operate end-to-end from a purely visual input, essentially also learning OCR in the process~\cite{kim22donut, lv2023kosmos}.

One of the most prominent recent developments in the field of AI has been the rise of large language models (LLMs) such as OpenAI's ChatGPT~\cite{liu23chatgpt}. These models have been found to excel at various natural language understanding tasks, and have been instruction-tuned to serve as open-domain problem solvers. Key to their success is their scale -- with large-scale training data and billions of parameters -- which leads to impressive capabilities~\cite{wei22emergent}. 
In contrast to the aforementioned multi-modal document comprehension models, traditional LLMs process only text\footnote{Though there is a recent trend towards multi-modal inputs, we will focus on large {\bf language} models in the strict sense here: The model's input and output are text sequences.}.
By that the modality of spatial layout, which seems vital for the processing of documents \cite{tang2023unifying, wang2023docllm}, is partially lost due to its reduction to a one-dimensional text sequence.

In this study we focus on an LLM-centric document comprehension pipeline that fuses the text with document layout.
First, a document's content is extracted with OCR, resulting in a set of words  equipped with box geometries. 
Second, this information is packaged into a purely textual representation that encodes both the document's text and its spatial structure. We will refer to this step as "verbalization" in the following.
Third, the resulting verbalized document is combined with the task description, resulting in a prompt for a pre-trained generative LLM, which solves the document comprehension task at hand without further fine-tuning.

This pipeline offers two {\it benefits}: First, it exploits the superior knowledge capacity and reasoning capabilities of LLMs -- which at the present time have been trained at larger scale and offer larger parametric capacity compared to current multi-modal document-specific models.
Second -- which is particularly relevant for practical applications -- the pipeline offers the benefit of simplicity, since it involves no model fine-tuning, thereby allowing us to keep a single generalist model.

Specifically, we focus on the key step of document verbalization, which raises several interesting questions: 
How well do LLMs perform at document comprehension tasks that involve challenging layout reasoning, even with no/little information on document geometry?
How are LLMs influenced by the way we feed them document representations and particularly, can we alter the document representations in a way that allows a LLM to exploit document geometry to achieve the same performance as a multi-modal model?

We investigate these questions with experiments on several document understanding datasets including tasks from the DUE benchmark, SROIE, WebSRC, and proprietary KIE datasets (from real-world industry scenarios). We examine two LLMs, namely ChatGPT3.5\footnote{gpt-3.5-turbo-1106} and the open-source LLM Solar\cite{kim2023solar}.

Overall, we make the following contributions:
\begin{enumerate}
\item A novel rule-based approach that enriches the prompts of existing text-centric LLMs with spatial structure information from documents.
The approach works across various kinds of documents and tasks and can be applied to various layouts without the need for fine-tuning.
\item A set of comprehensive experiments using both research and real-world document datasets as well as commercial and open-source models. We cover various document-specific tasks, different reading orders, and effects of noise being added to the OCR data.
\item Besides quantitative results, we also explore LLMs' limitations when it comes to interpreting  document layout in-depth on particularly challenging cases, for which we have annotated a subset of SROIE.
\footnote{Our annotated SROIE-Challenge dataset is available for future research, see Section~\ref{subsec:datasets}.}.
\item We also discuss efficiency issues, i.e. the extra tokens required for different approaches of encoding spatial layout information into the prompts.
 \end{enumerate}

\section{Related Work}

{\bf LLMs}: Language models built upon the attention-based transformer architecture~\cite{vaswani17transformer} are probably among the currently most intensely studied models in AI.
Due to the high growth they experienced, often involving several billion parameters, the capacity and reasoning capabilities of these models have rapidly progressed~\cite{wei22emergent}.
In addition to commercial providers such as OpenAI~\cite{gpt42023}, a variety of open-source models such as Llama 2~\cite{llama2} or Solar\cite{kim2023solar} are currently evolving. 
Two fundamental types of models are distinguished:
(1) Encoders, which generate representations of input texts and use them to can make decisions about texts. They are equipped with additional head layers that are be fine-tuned to the specific problem. 
(2) Decoders that generate text and can be instructed using prompts without additional training.
Recently, the latter paradigm has emerged as the dominant approach, as the resulting LLMs can serve as generalist agents for ad-hoc problem solving, without fine-tuning to specific tasks. 
Instruction tuning is used as an additional training step to facilitate this: 
It aims to bridge the gap between the LLM's goal of next-word prediction and the user's goal of having the LLM follow human instructions.~\cite{zhang2023instruction}.
Accordingly, we focus on instruction-tuned decoder models in this work.

{\bf Multi-Modal Models}: 
Many multi-modal models outsource OCR into preprocessing and operate on a combined input of document image and recognized text+geometry \cite{cao2022gmn,luo2023geolayoutlm,feng2023unidoc}: For example, the LayoutLM series, including the most recent version LayoutLMv3~\cite{layoutlm,layoutlmv2,layoutlmv3}, utilizes a BERT-type transformer encoder~\cite{devlin-etal-2019-bert}, which feeds on a concatenation of word embeddings and visual patch embeddings, and is trained with several masked language modeling (MLM) and word/patch alignment tasks. The model is applied to downstream tasks via fine-tuning specialied head models. Similarly, DocFormer~\cite{Appalaraju2021DocFormerET} applies an early fusion of image and text signals and a pre-training with global text-image alignment.
UDOP~\cite{Tang2022UnifyingVT} follows a generative approach and reconstructs text layout by an encoder-decoder model.

Other models operate end-to-end, feeding only on the document image and addressing text understanding in the process: Donut~\cite{kim22donut} uses an encoder-decoder architecture, which is pretrained to recognize the document images' text on large-scale real-world (IIT-CDIP) and synthetic documents.
Similarly, Dessurt~\cite{davis22dessurt} integrates OCR as part of its model. 
Many of the aforementioned papers include ablation studies demonstrating that models benefit from including geometry information in the input -- when trained accordingly. In this work, we extend this question to instruction-tuned LLMs. 

The work most similar to ours is LATIN-Prompt by Wang et al.~\cite{wang2023layout}, who have recently proposed a combination of a layout-aware document representation and a task-aware prompting, and have also investigated fine-tuning in the process. We extend on this work by (1) investigating multiple verbalization strategies, (2) thoroughly treating the prompt templates as a free, dataset-agnostic parameter to be optimized carefully and independently from the verbalization, and (3) explore the limitations of LLMs' layout reasoning capabilities in more detail by inspecting challenge cases and evaluating the effect of layout and OCR inaccuracies.

\section{Approach}
\begin{figure}
\centering
\includegraphics[width=\textwidth]{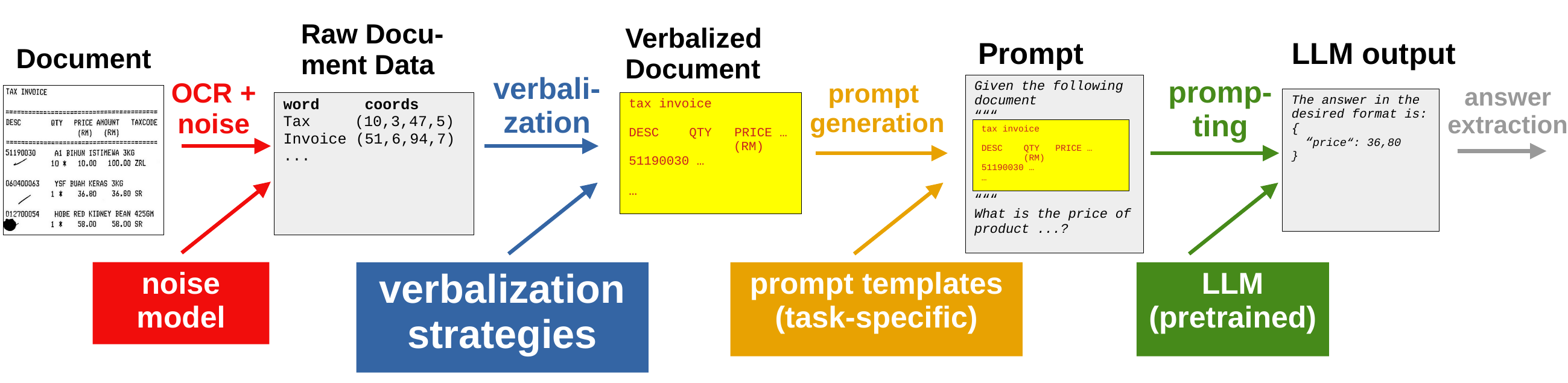}
\caption{Overview of our approach: Document OCR is converted into a text representation using different verbalization strategies (blue). Before verbalization, we optionally degrade the OCR by applying noise to the spatial position of OCR geometries (red).
The resulting document text representation is then inserted into a task specific prompt (yellow) and fed into a LLM (green).
Finally, the answers are extracted from the LLM output.
}
\label{fig:approach}
\end{figure}

Figure~\ref{fig:approach} shows an overview of our approach:
Given a document, we extract its text and corresponding word geometries using off the shelf OCR solutions.
The document is converted into a purely textual representation, using a step we refer to as \emph{verbalization}. We propose different verbalization strategies to add geometric and layout information to the textual document representation (see Section~\ref{subsec:verbalizers}).
To study the robustness of the verbalization with respect to inaccuracies of OCR geometries we degrade the OCR before verbalization by either applying noise to word positions or emulating layout analysis errors.
The verbalized document is then inserted into a 
prompt template together with task-specific directives, e.g. questions to be answered (Section~\ref{subsec:prompts}).
The prepared prompt is then fed into an LLM and the response is parsed from the output.

\subsection{Verbalizers} \label{subsec:verbalizers}
\begin{figure}
\centering
\includegraphics[width=1\textwidth]{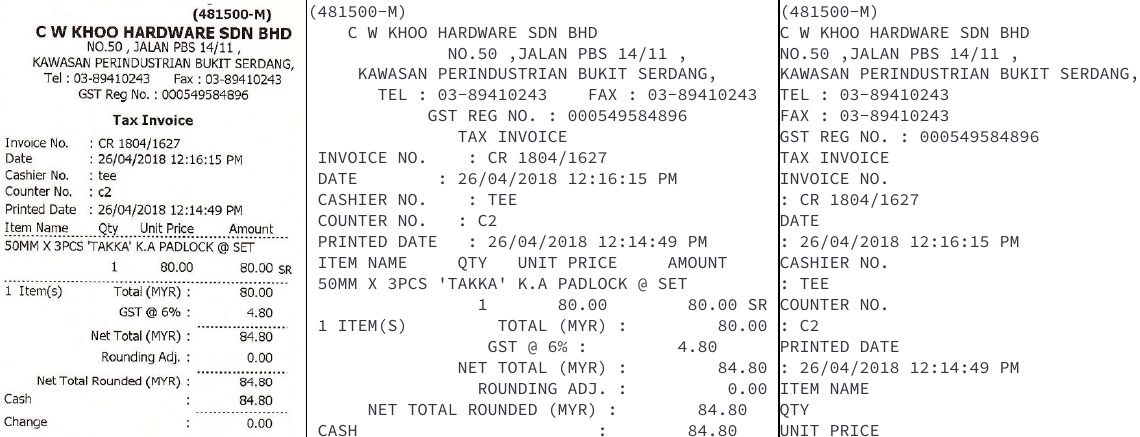}
\caption{Verbalization strategies on a random sample from the SROIE datset: original (left), \verbAF (middle) and \verbLU (right).} \label{fig:ascii_verb_example}
\end{figure}

We refer to verbalizers as strategies that create a textual document representation from an ordered collection of bounding boxes and the text associated with these boxes. 
This representation can serve as input to a text-based LLM.
Further, each verbalizer offers a textual description of its output format to guide the LLM in interpreting the verbalizer's outputs.
We outline multiple different verbalization strategies in the following. For each, we include the verbalization of an example word box with the text "TAX INVOICE" and 
coordinates $(x_{left},y_{top},x_{right},y_{bottom}) = (100, 50, 321, 100)$ and center point $(x,y) = (211, 75)$:

\begin{enumerate}
    \item \textbf{\verbLU} Serves as a baseline by only adopting the text \texttt{$t$} without extra layout information.
        The text lines retrieved from the OCR are concatenated using newlines to form the document representation. When no line candidates are available, we concatenate words with spaces.
        
    \item \textbf{\verbBB} Uses both the bounding box coordinates and the text.
        The box geometries are encoded together with the text of each box using a custom format.
        Coordinates are rounded to whole numbers and are encoded as "left", "top", "right" and "bottom".
        Example: \\
        \texttt{left:$100$ top:$50$ right:$321$ bottom:$100$ text:'$\text{TAX INVOICE}$'}
        
    \item \textbf{\verbBBM} Formats the bounding box coordinates in a XML style markup format followed by the text.
        Coordinates are rounded to whole numbers and are encoded as "left", "top", "right" and "bottom".
        Example: \texttt{\textless box left=$100$ top=$50$ right=$321$ bottom=$100$/\textgreater TAX INVOICE}
        
    \item \textbf{\verbCP} Formats both the bounding box center point coordinates in a XML style markup format followed by the text.
        Coordinates are rounded to whole numbers.
        Example: \\
    \texttt{\textless box x=$211$ y=$75$/\textgreater TAX INVOICE}
        
    \item \textbf{\verbAF} Uses the geometries to restore the original document layout via insertion of spaces and newlines. 
        To this end, the characters are placed on a grid such that their spatial location is similar to that on the document.
        Figure~\ref{fig:ascii_verb_example} shows a example output of the \verbAF verbalizer.
        At most 4 consecutive newlines are inserted.
        
    \item \textbf{\verbAFY} Similar to \verbAF, but it only encodes spatial information on the vertical dimension, i.e. only newlines are inserted and no spaces are used for horizontal alignment.
        At most 4 consecutive newlines are inserted.
     
    \item \textbf{\verbPH} Serves as a control run for the WebSRC dataset, where a structured  HTML representation of the document is available.
    Example: \\
        \texttt{\ldots\textless h3 tid="3"\textgreater TAX INVOICE\textless /h3\textgreater\ldots}
\end{enumerate}
When verbalizing with \verbAF and \verbAFY, 
each page is verbalized individually.
The resulting page verbalizations are then concatenated with an empty newline.

\subsection{Noise Models}\label{subsec:noise}
The OCR geometries generated by common OCR systems are subject to fluctuations and are rarely perfectly aligned with each other.
To study the robustness of the verbalization strategies with respect to inaccuracies in the order and spatial relationship of layout elements, we optionally apply noise models to the OCR output before feeding it into the verbalizers.
Each noising model takes an ordered list of bounding boxes as input, where the initial order corresponds to the reading order of the underlying OCR engine.

\begin{enumerate}
    \item \noiseNONE: Identity function. The coordinates and text of the OCR are not modified   (serves as a control run). 
    \item \noiseBBOXES: Degenerates each bounding box $b_i$ according to the formula $(x_0, y_0, x_2, y_2) \rightarrow (x_0{+}\Delta_i^x, y_0{+}\Delta_i^y, x_2{+}\Delta_i^x, y_2{+}\Delta_i^y)$, where $\Delta_i^x,\Delta_i^y \in [-20, 20]$ are uniformly sampled random numbers per box.
    Note that 20px is approximately the average character width in our data, such that two boxes can move up to 40px (or two letters) relative to each other.
    \item \noiseSHUFFLE: Shuffles the list of bounding boxes randomly.
    \item \noiseSEGMENT: Reorders a list of bounding boxes by selecting for each bounding box a successor box which is closer than \texttt{min\_char\_height} and \texttt{min\_char\_width} pixels. When there are no or multiple candidates, the successor box is selected  under consideration of the original order of the boxes. The procedure emulates the \emph{natural} reading order mode of Microsoft OCR\footnote{This behaviour of \emph{natural} reading order mode of Microsoft OCR has been shown empirically through our experiments.} and tends to read tables column wise instead of row wise, as the spacing between conse\-cutive rows is usually smaller than between consecutive columns.
\end{enumerate}

\subsection{Prompts}
\label{subsec:prompts}
To prompt the LLM, we insert the verbalized document into a task-specific prompt template.
As this prompt template influences quality just like the verbalization (order, wording and phrasing appear to matter), we aim to rigorously separate the effects of prompting from the effects of verbalization.
To determine a suitable prompt structure, we subdivide each prompt into common building blocks and determine an optimal composition.
Following known guidelines for prompt creation~\cite{OpenAIPrompting}, we generate 10 different prompt structures and evaluate them on a small subset of our data.
These structures differ in their ordering of the individual building blocks and are evaluated using a QA task on the SROIE Challenge dataset (see Section~\ref{subsec:datasets}).

\begin{figure}
\includegraphics[width=1\textwidth]{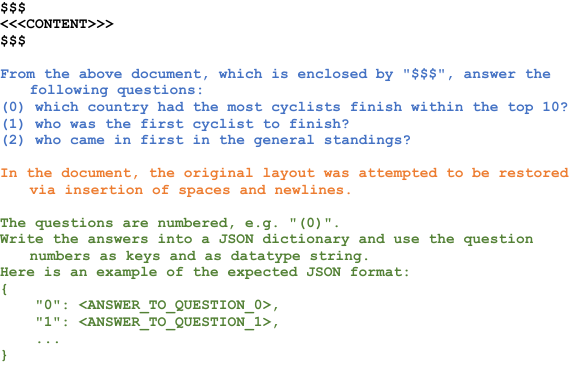}
\caption{Example for QA prompt \textbf{B} with three questions. Pattern \textbf{B} structure is: \texttt{DOCUMENT} (black), \texttt{TASK} (blue), \texttt{FORMAT} (orange) and \texttt{OUTPUT} (green).} \label{fig:prompt_example}
\end{figure}

Our prompts are divided into four components:
\texttt{DOCUMENT} corresponds to the verbalized document representation.
\texttt{TASK} encodes the task to solve.
\texttt{FORMAT} describes the format used for verbalization.
Finally, \texttt{OUTPUT} describes the expected output format using an example.
We identified two patterns \textbf{A},\,\textbf{B} to work best: \texttt{DOCUMENT TASK OUTPUT} (pattern \textbf{A}) and \texttt{DOCUMENT TASK FORMAT OUTPUT} (pattern \textbf{B}).
Based on these two structures, we create prompt templates for the tasks KIE, QA and NLI.
For efficency reasons, we group multiple questions (QA), statements (NLI) or keys which are to be retrieved (KIE) into a single prompt.
For QA and NLI samples that contain multiple questions to be answered, we enumerate those starting with 0.
Figure \ref{fig:prompt_example} shows an example for QA prompt \textbf{B} with multiple questions.\footnote{Please refer to Appendix~\ref{sec:appendix_prompts} for a comprehensive overview of the prompt templates}.

\subsection{Answer Extraction}
Due to the probabilistic nature of LLMs as text generators, their outputs are not guaranteed to conform with the requested format.
To ensure good readout of the answers, we process the output as follows:
\begin{enumerate*}[label=(\arabic*)]
    \item We request a single JSON object which assigns the answers to the respective enumeration numbers (QA, NLI) or keys (KIE).
    \item Given a single valid response object we parse the answers for the questions.
    \item Given multiple valid response objects we choose the object with the most answers for the questions asked.
    \item We use the enumeration number (QA, NLI) or the key (KIE) to extract a specific answer from the selected object.
    \item If no valid JSON object is returned, we do not generate any answer.
\end{enumerate*}
See Figure \ref{fig:prompt_example} for an example of the output format specification. 

While the generation of JSON works reliably in most cases, LLMs will occasionally generate output that does not parse to valid JSON objects.
Edge cases that we observed during development involve JSON objects which contain the correct value but a hallucinated key, e.g. \texttt{price\_of\_green\_tea} instead of \texttt{answer}.
Another common mistakes are nested objects, e.g. \texttt{\{"price": \{"green\_tea": ...\} \}}.
In these cases no answer is extracted.

\section{Experiments}
In the following experiments, we investigate whether suitable verbalization strategies can support LLMs with better layout reasoning and provide exemplary comparisons of open-source and commercial solutions.
In most experiments, we measure the awareness of the LLM towards layout aspects 
via  accuracy on document understanding tasks (which include research benchmarks and industry datasets, see Section~\ref{subsec:datasets}).
To investigate layout awareness in depth, we also take a qualitative look at a subset of manually annotated challenge cases (see Section~\ref{subsubsec:sroiequalanalysis}).

\subsection{Datasets}
\label{subsec:datasets}

{\bf DUE Benchmark}
We evaluate our approach using the DUE benchmark\cite{Borchmann2021DUEED}, 
specifically on the datasets \emph{DocVQA}, \emph{InfographicsVQA}, \emph{TabFact} and \emph{WikiTableQuestions} with the tasks VQA (DocVQA \& InfographicsVQA), TableNLI (TabFact) and TableQA (WikiTableQuestions).
We did not analyze the other datasets DeepForm, Kleister Charity and PWC, which are also part of the due benchmark, as these documents have a very high number of pages\footnote{A limitation when working with long documents is the context length of LLMs. While solutions to this exist, such verbose documents are not part of our scope.}.

{\bf WebSRC}
WebSRC\cite{chen-etal-2021-websrc} is a collection of 
360K question-answer pairs, which are collected from 60 different websites spanning 11 different domains.
Besides the QA pairs the dataset also consists of web page segments, where each consists of a simplified version of the source HTML, a screenshot and a JSON file which contains additional spatial and layout information.
Due to difficulties in retrieving text level bounding boxes from the JSON and HTML data, we manually perform OCR on the screenshots and use this data for further evaluation.\footnote{This OCR data has been contributed to the authors of WebSRC and is also made publicly available at \url{https://github.com/46692/WebSRC_OCR}}
Only this dataset uses the \verbPH verbalizer with the given HTML.

{\bf SROIE and SROIE Challenge}
SROIE~\cite{SROIE} is a collection of 973 scanned receipts and the corresponding OCR results.\footnote{We use the revised version of the dataset from \url{https://www.kaggle.com/datasets/urbikn/sroie-datasetv2}}
The task of the dataset is KIE with 4 keys to be extracted for each sample: company, date, address and total.

The original SROIE asks for the same 4 keys to be extracted for each sample.  We argue that these keys in particular require no comprehensive understanding of the document's layout.
For example, the company name is almost always the first thing written on the receipt, where date and address follow shortly after.
To investigate LLM's understanding of layout more closely, we created a challenge set that queries the value of a specific table cell (\emph{"How many of the item 'Green Tea' were purchased?"}) or directly reference the document's layout (\emph{"Which entity is written above the card expiry date?"}).
To do so, we manually annotated 101 samples from the train split of the SROIE dataset to create a challenging QA dataset\footnote{Made publicly available at \url{https://github.com/46692/SROIEChallenge}}.
We categorize the questions into \emph{quantity}, \emph{currency} and \emph{string}, where the latter refers to any other question that corresponds to neither of the first two types.

{\bf Proprietary KIE Datasets}
We further evaluate KIE performance on two proprietary KIE datasets from Insiders Technologies, which both contain particularly diverse and challenging examples from real world business correspondence:
\textbf{ITForms} is a collection of 100 multipage form documents in German language, with 9 keys each.
It includes, among others, forms for applying for insurance benefits, registering vehicles and bank forms, e.g. opening a depot.
\textbf{ITInvoices} is a collection of 104 invoice documents in German language, which are predominantly single page and contain 21 keys each.
It includes both business invoices as well as scanned receipts.

\subsection{Setup}
\label{subsec:setup}

{\bf LLMs}
We evaluate our approach with two LLMs: ChatGPT and Solar\cite{kim2023solar}.
For the evaluation of ChatGPT we use \texttt{gpt-3.5-turbo-1106}\footnote{The model was used in the period of November 2023 to February 2024.} in JSON mode\cite{OpenAIJsonMode}, with a temperature of $0$, and enter each prompt in the role of \texttt{user}.
We further evaluate the 8 bit quantized version\footnote{\url{https://huggingface.co/upstage/SOLAR-0-70b-8bit}} of the recent open-source LLM Solar 70b on SROIE and SROIE Challenge.\footnote{The evaluation of other datasets had to be omitted due to time constraints.}
For Solar each prompt is also entered in the role of \texttt{user}.\footnote{As stated on the Hugging Face model card, Solar expects the role being given in the prompt. 
Therefore we used the following wrapper \texttt{\#\#\# User:PROMPT\textbackslash n\textbackslash n\textbackslash n\#\#\# Assistant:} where \texttt{PROMPT} is replaced with the prompt prepared by our pipeline and \texttt{\textbackslash n} symbolizes an empty line.}
Unless stated otherwise, experiments use the ChatGPT model and prompt template \textbf{A}, i.e. the template without verbalizer format description.

{\bf OCR}
Each dataset in the DUE benchmark comes with a selection of pre-applied OCR engines, where we use \texttt{microsoft\_cv} and \texttt{tesseract} as a fallback in case the former is not available, which is only the case for TabFact.
The OCR results in these datasets contain information about the page and line index, which is used to join all word bounding boxes on the same page with the same line index together.
Microsoft Computer Vision OCR is used for WebSRC, ITForms and ITInvoices.
We contribute the OCR for WebSRC train and test splits.
For SROIE and SROIE Challenge we use the OCR results delivered with the dataset.

{\bf Metrics}
For evaluation of the DUE datasets we use the official evaluation framework\footnote{\url{https://github.com/due-benchmark/evaluator}} with the metrics given in \cite{Borchmann2021DUEED}: ANLS for DocVQA and InfographicsVQA and accuracy for TabFact and WikiTableQuestions.
WebSRC is evaluated according to the procedure in the GitHub repository\footnote{\url{https://github.com/X-LANCE/WebSRC-Baseline}} and the scores are given as EM and F1.
For SROIE, SROIE Challenge, ITForms and ITInvoices we create a type aware accuracy measure:
Each response is assigned one of four types based on the expected response, which describe how it is compared to the ground truth (the procedures described are applied to both the GT and the response extracted from the LLM output):
For \texttt{string} values a case-insensitive comparison is made. 
\texttt{date} values are parsed via the Python dateparser library\footnote{\url{https://github.com/scrapinghub/dateparser}} and then compared for equality.
\texttt{currency} values are sanitized via a RegEx\footnote{\texttt{\textbackslash d+(?:(\textbackslash.|,)\textbackslash d{1,2})?}}, replacement of commas with dots and then compared for equality.
\texttt{quantity} values are sanitized via a RegEx\footnote{\texttt{(?:[ a-zA-Z]*)(\textbackslash d+)(?:[ a-zA-Z]*)}} and then compared for equality.
The proposed accuracy measure defaults to EM for \texttt{string} and also for \texttt{currency} and \texttt{quantity} after units are neglected, i.e. no rounding is performed for the latter two. In case a value cannot be parsed to its specified type, an empty answer is returned.

\subsection{Results}
See Section~\ref{sec:token_count_analysis} in the appendix for a comparison of the token overhead added by each verbalization strategy.
See Section \ref{sec:format_description_analysis} in the appendix for an analysis of the effects that the verbalizer format description has on the different prompt templates $A$ and $B$.

\subsubsection{Dataset Results}

We report the results on the various datasets with the metrics laid out in section~\ref{subsec:setup}: type aware accuracy for SROIE, SROIE Challenge, ITForms, ITInvoices; ANLS for DocVQA, InfographicsVQA; accuracy for TabFact, WikiTableQuestions; F1 and EM for WebSRC.
The results in tables~\ref{tab:dataset_results_due} and \ref{tab:dataset_results_other} show, the our approach can compete with state-of-the-art models.
Specifically, the results of the DUE benchmark in table~\ref{tab:dataset_results_due} demonstrate that the introduction of layout information to the prompt proves beneficial.
Our approach achieves state-of-the-art performance on InfographicsVQA and WikiTableQuestions.\footnote{State as of 10th February 2024 according to \url{https://duebenchmark.com}}
Throughout the benchmark, \verbAF proves to be the best verbalization strategy on average. With a peak gain of of 15\% (from 47.7\% to 54.9\%) on InfoVQA.

Table~\ref{tab:dataset_results_other} shows that we achieve competitive results for some of the other datasets. 
Specifically, our approach achieves the 3rd best F1 score on WebSRC with the \verbAF verbalization.\footnote{State as of 10th February 2024 according to~\cite{WebSRCLeaderboard}}
The \verbPH baselines further shows promising results for HTML formatted document representations, achieving best performance out of all verbalizations.
However, this verbalization strategy is not viable for real world documents, as these would have to exist as HTML documents in the first place or would introduce a separate layout processing model into the pipeline, eliminating the need for our approach.
Results on SROIE show that $\text{StructTexT}$ significantly outperforms our approach, demonstrating the superiority of multi-modal models on the dataset.
Comparison with the other datasets is difficult:
For our custom SROIE Challenge subset no comparisons exist of course.
While ITForms and ITInvoices are proprietary datasets that do not allow direct comparison with other approaches, they give us an insight into how the models operate on real world business documents.
These data sets are characterized by the fact that not all keys have to generate a value.
Information is often missing on real documents and not every key can be assigned to a value.
The model must therefore have sufficient ability to reject a value, i.e. it should only output a value if it can be found on the document.
Our results show that the LLM-based approaches perform significantly worse on these datasets.
We observed that in most cases the LLMs produce outputs and rarely provide an empty response, which lowers their overall score in the evaluation.
We believe that this problem can be reduced by clearer instructions in the prompt.

A trend that can again be observed throughout the datasets is the good performance of \verbAF and \verbAFY among the verbalization strategies.

\begin{table}[]
	\centering
    \caption{Comparison with other models published on the DUE-Benchmark. Underlines denote the best verbalization strategy in the dataset.
    It shows that our approach achieves competitive results and even state-of-the-art results on the two datasets InfographicsVQA and WikiTableQuestions, with an improvement of 15\% compared to the baseline for the former.
    On WebSRC, we rank third in terms of F1 score.
    Further, it is shown that \verbAF is the best verbalization strategy among the ones tested. 
    *LATIN Prompt templates appear to be optimized for the specific dataset and partially include excerpts of examples.}
    \label{tab:dataset_results_due}
	\begin{tabular}{@{}llccclccc@{}}
	\toprule
	\multicolumn{2}{l}{\multirow{2}{*}{Model}} & \multirow{2}{*}{Modality} & \multicolumn{2}{c}{Question Answering} &  & \multicolumn{2}{c}{Table QA/NLI} & \multirow{2}{*}{Avg.} \\ \cmidrule(lr){4-5} \cmidrule(lr){7-8}
	\multicolumn{2}{l}{}                       &                           & DocVQA            & InfoVQA            &  & WTQ           & TabFact          &                       \\ \midrule
   	\multicolumn{2}{l}{$\text{BERT}_{\textsc{LARGE}}$~\cite{devlin-etal-2019-bert}}  & T          & 67.5                   & -                   &  & -              & -                 & -                      \\
	\multicolumn{2}{l}{Donut~\cite{kim22donut}}                  & V                         & 72.1              & -                  &  & -             & -                & -                     \\
	\multicolumn{2}{l}{$\text{T5}_{\textsc{LARGE}}$+2D+U~\cite{powalski2021going}}  & T+L          & 81.0                   & 46.1                  &  & 43.3              & 78.6                 & 62.3                      \\
	\multicolumn{2}{l}{$\text{LayoutLMv2}_{\textsc{LARGE}}$ + QG~\cite{layoutlmv2}}  & T+L+V          & \textbf{86.7}                   & -                   &  & -              & -                 & -                      \\
 	\multicolumn{2}{l}{$\text{LayoutLMv3}_{\textsc{LARGE}}$~\cite{layoutlmv3}}  & T+L+V          & 83.4                   & 45.1                   &  & 45.7              & 78.1                 & 63.1                      \\
 	\multicolumn{2}{l}{UDOP~\cite{tang2023unifying}}  & T+L+V          & 84.7                   & 47.4                   &  & 47.2              & \textbf{78.9}                 & \textbf{64.6}                      \\
\multicolumn{2}{l}{LATIN-Prompt (Claude)~\cite{wang2023layout}}  & T+L          & 82.6                   & \emph{54.5}*                    &  & -              & -                 & -                      \\
  \hline
	Ours            & \verbLU            & T                         & 76.3          & 47.7           &  & 45.1          & 68.4            & 59.4                      \\
	Ours            & \verbAF           & T+L                       & \underline{79.8}           & \textbf{\underline{54.9}}  &  & \textbf{\underline{47.7}} & 70.1& \underline{63.1}          \\
        Ours            & \verbAFY          & T+L                       & 76.3           & 49.6           &  & 45.5          & \underline{70.3}& 60.4                      \\
        Ours            & \verbBB              & T+L                       & 74.8        & 46.4           &  & 35.0          & 68.5            & 56.2                      \\
        Ours            & \verbBBM        & T+L                       & 74.6             & 45.8           &  & 36.2          & 68.6            & 56.3                      \\
        Ours            & \verbCP              & T+L                       & 75.1        & 47.4           &  & 38.2          & 67.8            & 57.1                      \\\bottomrule
	\end{tabular}
\end{table}


\begin{table}[]
	\centering
    \caption{Evaluation results for SROIE, ITForms, ITInvoices, WebSRC and SROIEChallenge. Underlines denote the best verbalization strategy for the dataset. WebSRC results of other models are taken from the official leaderboard and show that the performance our approach is close to that of the third-placed model. Proprietary KIE Model refers to an internal model of Insiders Technologies, which is a multi-modal LLM free approach.: ITForms and ITInvoices contain samples for which not all keys have a value on the documents. While this works to some extent, it is not properly supported using our current prompt. *For WebSRC left score is EM and right score is F1.}
    \label{tab:dataset_results_other}
 \resizebox{1\textwidth}{!}{
	\begin{tabular}{@{}lcccclcc@{}}
	\toprule
	\multirow{2}{*}{Model} & \multirow{2}{*}{Modality} & \multicolumn{3}{c}{KIE} &  & \multicolumn{2}{c}{Question Answering} \\ 
 \cmidrule(lr){3-5} \cmidrule(lr){7-8}
            &       & SROIE &ITForms& ITInvoices &                    & WebSRC*        & SROIEChallenge  \\ \midrule
SageGPT-small-v0.2~\cite{WebSRCLeaderboard}            & ?     & -     & -     & -             && \textbf{89.1 / 92.2}  & -  \\
DocPrompt (ErnieLayout-Large)~\cite{wu2023docprompt} & T+L+V & -     & -     & -             && 77.4 / 85.0  & -  \\
TIE (MarkupLM-Large)~\cite{li2021markuplm}          & T+L   & -     & -     & -             && 76.3 / 80.5  & -  \\
$\text{StructTexT}$~\cite{li2021structext} & T+L+V & \textbf{98.7} & - & - & - & - \\
Proprietary KIE Model         & T+L & 91.7     & \textbf{86.2}  & \textbf{90.1}  && -            & -     \\
\hline
\verbLU                  & T     & \underline{79.9}  & 68.4  & 54.5          && 72.9 / 80.5  & 81.2  \\
\verbAF                  & T+L   & 77.0  & \underline{73.9}  & 54.2          && 74.2 / 80.7  & \textbf{\underline{86.1}}  \\
\verbAFY                 & T+L   & 79.0  & 69.0  & \underline{54.6}          && 72.4 / 80.3  & 81.2  \\
\verbBB                  & T+L   & 75.4  & 64.2  & 54.1                      && 68.3 / 76.6  & 72.3  \\
\verbBBM                 & T+L   & 74.3  & 65.6  & 53.9                      && 68.1 / 75.9  & 71.3  \\
\verbCP                  & T+L   & 73.3  & 65.1  & 51.8                      && 68.9 / 76.9  & 74.3  \\
\verbPH                  & T+L   & -     & -     & -                         && \underline{80.0 / 84.1}  & -     \\\bottomrule
	\end{tabular}
 }
\end{table}


\subsubsection{Comparison of ChatGPT to Solar}
We compare the performance of our approach when applied to two different LLMs, specifically ChatGPT 3.5 and Solar70B8Bit.
The results in table~\ref{tab:chatgpt_vs_solar} show that open-source LLMs provide a viable alternative to commercial solutions for document comprehension using our approach:
The results of both LLMs are on par on SROIE, with Solar performing slightly better.
On SROIE Challenge, ChatGPT has a lead of 4.8 pp. on average.
Further, it is shown that Solar is apparently able to make better usage of the layout information delivered by \verbBB, \verbBBM and \verbCP verbalizers compared to ChatGPT.

While it is unclear whether SROIE is part of ChatGPT's training data, we checked the training data of Solar\footnote{As given under \url{https://huggingface.co/upstage/SOLAR-0-70b-8bit}} and could find no SROIE data.
However, we can assure that neither of both models has seen the questions of SROIE Challenge during training.

\begin{table}[]
	\centering
    \caption{Evaluation results for the comparison of Solar and ChatGPT 3.5. Underlines denote the best verbalization strategy for the LLM in the dataset. It shows that open-source LLMs provide a viable alternative to commercial solutions for docu\-ment comprehension using our approach: Solar performs slightly better than ChatGPT on SROIE, while ChatGPT has an advantage of 4.8 pp. on our custom SROIE Challenge set.}
    \label{tab:chatgpt_vs_solar}
	\begin{tabular}{@{}llccclcc@{}}
	\toprule
	\multicolumn{2}{l}{\multirow{2}{*}{Verbalizer}} & \multirow{2}{*}{Modality} & \multicolumn{2}{c}{SROIE} &  & \multicolumn{2}{c}{SROIE Challenge} \\ \cmidrule(lr){4-5} \cmidrule(lr){7-8}
	\multicolumn{2}{l}{}                       &                           & ChatGPT & Solar              &  & ChatGPT  & Solar \\ \midrule
                & \verbLU         & T                         & \underline{79.9}  & 76.6   &  & 81.2 & 72.3       \\
                & \verbAF         & T+L                       & 77.0  & 76.7   &  & \underline{86.1} & \underline{81.2}        \\
                & \verbAFY        & T+L                       & 79.0  & 77.3   &  & 81.2 & 79.2        \\
                & \verbBB         & T+L                       & 75.4  & 76.2   &  & 72.3 & 66.3        \\
                & \verbBBM        & T+L                       & 74.3  & \underline{77.7}   &  & 71.3 & 66.3        \\
                & \verbCP         & T+L                       & 73.3  & 76.9   &  & 74.3 & 72.3        \\\midrule
                \multicolumn{2}{l}{Avg.}       &              & 76.5  & \textbf{76.9}      &  & \textbf{77.7} & 72.9 \\\bottomrule
	\end{tabular}
\end{table}

\subsubsection{Noise Model Analysis}
\begin{figure}
\centering
\includegraphics[width=1\textwidth]{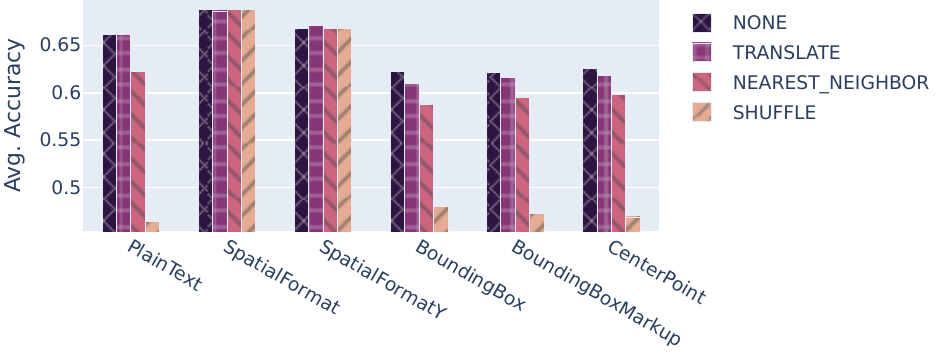}
\caption{Comparison of the noise models for each verbalizer.
Values shown are scores averaged over all datasets.
It shows, that \verbLU's performance diminishes when the layout is misinterpreted. 
Further is shown that \verbAF and \verbAFY verbalizers are the least prone to noise introduced to the OCR data.
Note that they are not affected by changes to the bounding box ordering, as they operate only using the bounding box coordinates.}
\label{fig:noise_all_datasets}
\end{figure}

We evaluate the robustness of our verbalization strategies against noise and layout misinterpretations introduced to the document data. We simulate this by applying the noise models \noiseBBOXES, \noiseSHUFFLE and \noiseSEGMENT (see Section~\ref{subsec:noise}).
For each noise model, the average of the scores achieved with each verbalization strategy across all datasets is determined.
\footnote{In line with the DUE benchmark, we resort to an arithmetic mean of 
different metrics.~\cite{Borchmann2021DUEED} For WebSRC, where two metrics are reported, we use the F1 score.}
The results presented in figure~\ref{fig:noise_all_datasets} show that \verbAF and \verbAFY are the least affected by the noise models.
Further, it shows that \verbLU is very susceptible to wrong layout interpretation by the OCR system.

\subsubsection{Qualitative Analysis: SROIE Challenge}\label{subsubsec:sroiequalanalysis}
We explore the impact of document layout on LLMs in-depth on our SROIE Challenge dataset, which features demanding questions on specific table cells and the relative position of items.
See Section~\ref{sec:sroie_qualitative_analysis_appendix} in the appendix for  examples of these challenge cases.
We found the LLM to work surprisingly well, answering 59 of 101 questions with all verbalizers and 82 with the \verbLU verbalizer.
The failures on the remaining 19 samples were related to layout misinterpretations that follow directly from the limited plain-text verbalization: 
\begin{enumerate*}[label=(\roman*)]
    \item column-wise order of OCR output instead of a row-wise 
    \item delayed table cells, which were placed after all other cells at the end of a table. 
\item overly complex samples (e.g. tables spanning over 12 rows and 6 columns)  
\item empty cells, which lead the LLM to wrong conclusions based on the ordering of the bounding boxes, and 
\item neighboring cells merged to a single bounding box.
\end{enumerate*}
Especially combinations of these factors provided challenging cases.
Overall, the challenge cases were more reliably solved by the \verbAF 
strategy (87 out of 101 correct).
This indicates -- on a limited number of manually inspected samples -- a better resiliency of the \verbAF verbalization with respect to OCR layout misinterpretations.

\section{Conclusion}
We have investigated techniques for adding layout information to prompts for instruction-tuned LLMs to enhance  document understanding performance.
This approach only requires pre-processing of document text and the prompt without the need for extra fine-tuning.
We achieve higher scores compared to layout-unaware document representations on 7 out of 9 datasets across different document tasks 
, reaching state-of-the-art results on two datasets and often 
times yielding results competitive with those of specially trained multi-modal models. 
We have shown that our approach works for both commercial as well as open-source LLMs.
A potential threat to validity is that that standard datasets may have been part of the training data for the proprietary LLM ChatGPT, while this has been shown not to be the case for Solar.
Our results indicate, however, that the improvements of our approach on the non-public datasets (ITForms, ITInvoices) and the   specifically annotated (SROIE-Challenge), are in line with the findings on public datasets.
The proposed method is particularly suited for structured documents that make heavy use of spatial alignments and blanks.
In these cases, our approach should be considered as the best model choice between text-based LLMs and multi-modal document transformers.
It comes with little to no overhead and requires no extra training while also not adding significantly more tokens to the input.
For future research, an interesting study focus are recent instruction-tuned LLMs with additional visual input such as GPT-4~\cite{ChatGPT4MultiModal}. 
As these models are associated with higher costs, we have focused on text-only representations in this work. However, exploring the benefits and trade-offs of including visual input is worthwhile.
We also recon that more work is needed when scaling our solution to multi-page reasoning problems, especially when the number of pages becomes larger.

\appendix
\section{Prompt Templates}\label{sec:appendix_prompts}

For each task QA, NLI and KIE 2 prompt templates \textbf{A} and \textbf{B} are created.
This section gives an overview of the prompt templates used.

\begin{tabular}{ccc}

\begin{minipage}{.5\textwidth}
\begin{lstlisting}[caption={QA prompt template \textbf{B}. In version \textbf{A} of the template \placeholder{FORMAT} is removed.},captionpos=b,breaklines=true,basicstyle=\tiny,tabsize=2]
$$$
<<<CONTENT>>>
$$$

From the above document, which is enclosed by "$$$", answer the following questions:
<<<QUESTION>>>

<<<FORMAT>>>

The questions are numbered, e.g. "(0)".
Write the answers into a JSON dictionary and use the question numbers as keys and as datatype string. 
Here is an example of the expected JSON format:
{
    "0": <ANSWER_TO_QUESTION_0>,
    "1": <ANSWER_TO_QUESTION_1>,
    ...
}
\end{lstlisting}
\end{minipage}

&

\hspace{0.1cm}

&

\begin{minipage}{.525\textwidth}
\begin{lstlisting}[caption={NLI prompt template \textbf{B}. In version \textbf{A} of the template \placeholder{FORMAT} is removed.},captionpos=b,breaklines=true,basicstyle=\tiny,tabsize=2]
$$$
<<<CONTENT>>>
$$$

From the above document, which is enclosed by "$$$", validate the following statements:
<<<QUESTION>>>

<<<FORMAT>>>

The statements are numbered, e.g. "(0)".
Write the answers into a JSON dictionary and use the statement numbers as keys and as datatype string. 
Answer with the string value "1" in case of a true statement and with the string value "0" in case of a false statement.
Here is an example of the expected JSON format:
{
    "0": <ANSWER_FOR_STATEMENT_0>
    "1": <ANSWER_FOR_STATEMENT_1>,
    ...
}
\end{lstlisting}
\end{minipage}

\\ 

\begin{minipage}{.525\textwidth}
\begin{lstlisting}[caption={KIE prompt template \textbf{B}. In version \textbf{A} of the template \placeholder{FORMAT} is removed.},captionpos=b,breaklines=true,basicstyle=\tiny]
$$$
<<<CONTENT>>>
$$$

From the above document, which is enclosed by "$$$", extract the values to the following keys:
<<<QUESTION>>>

<<<FORMAT>>>

Write the answers into a JSON dictionary with one entry for each requested key.
Here is an example of the expected JSON format:
{
    KEY0: <VALUE_FOR_KEY_0>,
    KEY1: <VALUE_FOR_KEY_1>,
    ...
}
\end{lstlisting}
\end{minipage}

&

\hspace{0.1cm}

&

\begin{minipage}{.525\textwidth}
\end{minipage}
\end{tabular}

\section{Verbalizer Token Count Analysis}\label{sec:token_count_analysis}
\begin{figure}
\centering
\includegraphics[width=1\textwidth]{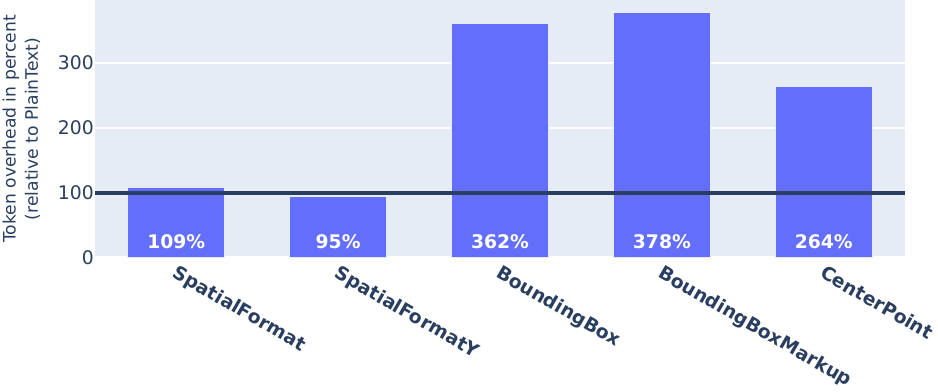}
\caption{Relative token overhead introduced by each verbalization strategy compared to \verbLU baseline. 
Values are given in percentage of number of tokens required by \verbLU verbalizer and averaged over all documents in all datasets. 
It shows that \verbAF and \verbAFY introduce the least token overhead compared to the other verbalizers. Furthermore, it shows that \verbAFY requires in even less tokens than the \verbLU baseline.} \label{fig:token_count_analysis}
\end{figure}

We quantify the additional tokens added to the prompt by each verbalization strategy.
To this end, we compare the number of tokens required by the \verbLU verbalizer to the number of tokens required by the other verbalizers. 
Tokens are counted with the tiktoken\footnote{\url{https://github.com/openai/tiktoken}} BPE tokenizer, which is provided by OpenAI for use with their models.
The analysis plotted in figure~\ref{fig:token_count_analysis} shows that \verbAF and \verbAFY verbalization strategies introduce the least token overhead compared to the \verbLU baseline.
Counterintuitively, it further shows that \verbAFY results in less tokens than the baseline.

\section{Format Description Analysis}\label{sec:format_description_analysis}
\begin{figure}
\centering
\includegraphics[width=1\textwidth]{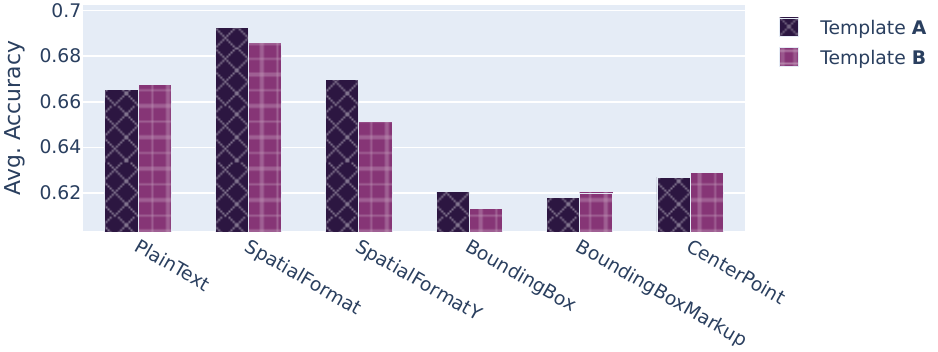}
\caption{Comparison of the impact of verbalizer format description in the prompt.
Values shown are scores averaged over the datasets SROIE, SROIE Challenge, DocVQA, InfographicsVQA, TabFact and WikiTableQuestions.
It shows that the addition of the format description lead to no significant changes, proving beneficial for some and detrimental for other verbalization strategies.
Further, the results were inconclusive and differed for each dataset: on some, the format description benefits some verbalizers, on other datasets it does not.}
\label{fig:format_description_analysis}
\end{figure}

We evaluate the benefit added by including a verbalizer-specific format description (see Sections~\ref{subsec:verbalizers} and~\ref{subsec:prompts}.
Figure~\ref{fig:format_description_analysis} shows that the format description did not lead to a significant performance gain, and that some verbalization strategies actually declined in performance.
Curiously, the results vary from dataset to dataset, e.g. \verbAF benefits on SROIE from the format description, but declines in performance on all other datasets.
The results are inconclusive and the performance changes never exceed 2 pp, which we see as insignificant.
In addition to that, only one format description for each verbalizer was evaluated, while a vast amount of possible format descriptions exist for each.

\section{SROIE Challenge Examples}\label{sec:sroie_qualitative_analysis_appendix}
\begin{figure}[h]
    \centering
       \subfloat[X51006619772]{%
          \includegraphics[height=10cm]{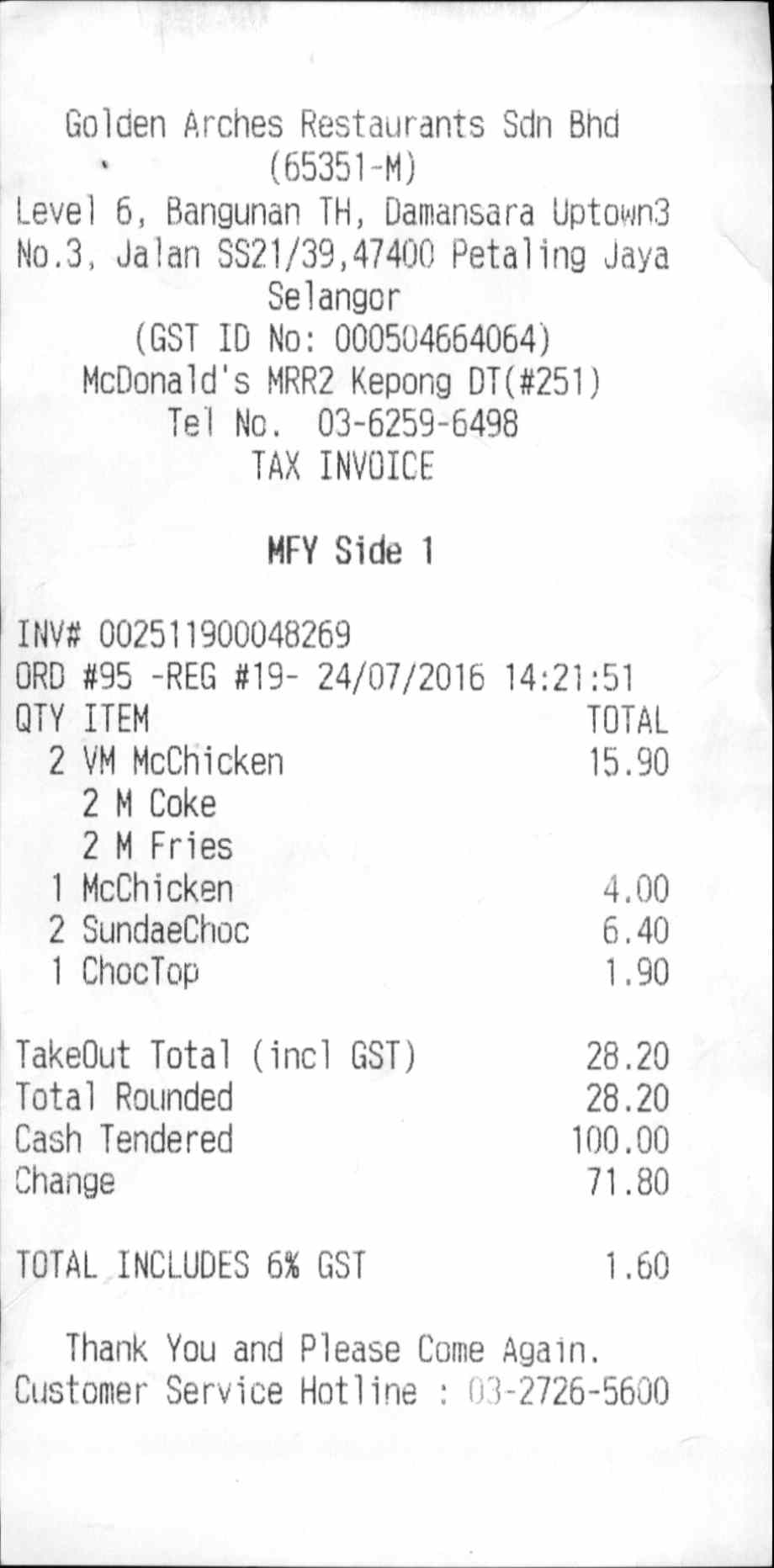}%
          \label{fig:left}%
       } 
       \subfloat[X51005806679]{%
          \includegraphics[height=10cm]{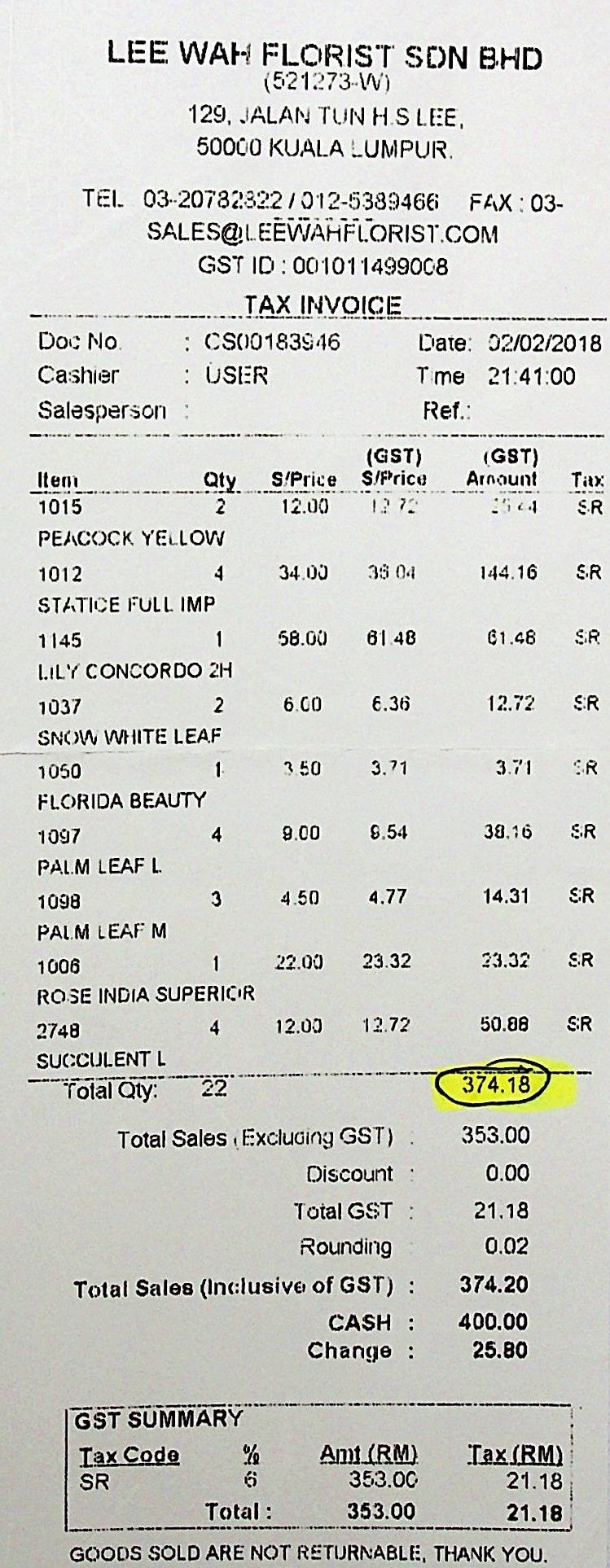}%
          \label{fig:middle}%
       }
       \subfloat[X51006332575]{%
          \includegraphics[height=10cm]{img/sroie_bboxmerge.jpg}%
          \label{fig:right}%
       }
       \caption{Overview of three difficult cases in the SROIE Challenge dataset: (a) shows a receipt with empty cells, which leads the LLM to wrong conclusions based on the ordering of the bounding boxes when using the \verbLU baseline.
       The receipt shown in (b) has a rich layout, with a table consisting of 6 columns and 18 rows.
       Due to layout misinterpretation of the OCR system, the texts \texttt{QTY} and \texttt{U/PRICE} were merged into a single bounding box in (c).}
       \label{fig:sroie_challenge_cases}
\end{figure}

Figure~\ref{fig:sroie_challenge_cases} shows three difficult cases from the SROIE Challenge dataset, for three of the categories mentioned in Section~\ref{subsubsec:sroiequalanalysis}:
(iii) overly complex samples, (iv) empty cells, which lead the
LLM to wrong conclusions based on the ordering of the bounding boxes, and (v) neighboring cells merged to a single bounding box.

\clearpage
\printbibliography

\end{document}